\documentclass[letterpaper]{article} 
\usepackage[preprint]{aaai2027}  
\usepackage[hyphens]{url}  
\usepackage{multirow}
\usepackage{graphicx} 
\usepackage{natbib}  
\usepackage{caption} 
\usepackage{algorithm}
\usepackage{algorithmic}
\usepackage{booktabs}
\usepackage{graphicx}
\usepackage[table]{xcolor}
\usepackage{amssymb}
\usepackage{placeins}
\usepackage{tabularx}
\usepackage{array}
\newcolumntype{Y}{>{\centering\arraybackslash}X}

\usepackage{newfloat}
\usepackage{listings}
\DeclareCaptionStyle{ruled}{labelfont=normalfont,labelsep=colon,strut=off} 
\floatstyle{ruled}
\newfloat{listing}{tb}{lst}{}
\floatname{listing}{Listing}

\usepackage{booktabs}

\usepackage{amsmath}
\usepackage{amssymb}

\title{GeoPrompt: Geometry-Conditioned Prompt Learning for Underwater Open-Vocabulary Instance Segmentation}

\author{
    Shanshan Han\textsuperscript{\rm 1},
    Yuxi Wang\textsuperscript{\rm 1}\corresponding,
    Can Qin\textsuperscript{\rm 1},
    Yujie Chen\textsuperscript{\rm 1},\\
    Cong Pan\textsuperscript{\rm 2},
    Haitao Li\textsuperscript{\rm 1},
    Guoqiang Zhong\textsuperscript{\rm 1}\corresponding
}

\affiliations{
    \textsuperscript{\rm 1}Ocean University of China, Qingdao, China\\
    \textsuperscript{\rm 2}Nanjing University of Aeronautics and Astronautics, Nanjing, China\\
    yuxi.wang@ouc.edu.cn
}

\begin{document}

\maketitle

\begin{abstract}
Open-vocabulary instance segmentation (OVIS) provides a promising solution for marine perception by recognizing and segmenting categories beyond predefined label spaces. Existing OVIS methods usually rely on vision-language alignment, where visual region features are matched with textual category embeddings to recognize both seen and unseen classes. However, this alignment assumption becomes fragile in underwater environments. Color attenuation, scattering, turbidity, and low contrast distort visual appearance, causing region features to deviate from their corresponding semantic concepts. As a result, the similarity between degraded visual features and textual classifiers becomes unreliable, which particularly weakens the recognition of unseen underwater categories that cannot be corrected by category-specific supervision. To address these issues, we propose GeoPrompt, a geometry-conditioned prompt learning framework for underwater open-vocabulary instance segmentation. Our key insight is that geometric structure is more stable than visual appearance in degraded underwater scenes. Specifically, GeoPrompt introduces a geometry-guided visual adaptation module~(GVAM) to extract depth-derived structural priors and inject them into multi-scale visual features. It further develops a geometry-conditioned prompt learning~(GCPL) that generates geometry-aware textual classifiers from structural representations. These geometry-aware classifiers are integrated with fixed and image-conditioned textual classifiers to improve open-vocabulary semantic alignment. Experiments on the MARIS-OpenIS benchmark show that GeoPrompt achieves state-of-the-art performance and brings consistent improvements on underwater open-vocabulary instance segmentation.
\end{abstract}

\begin{figure*}[t]
    \centering
    \includegraphics[width=0.95\textwidth]{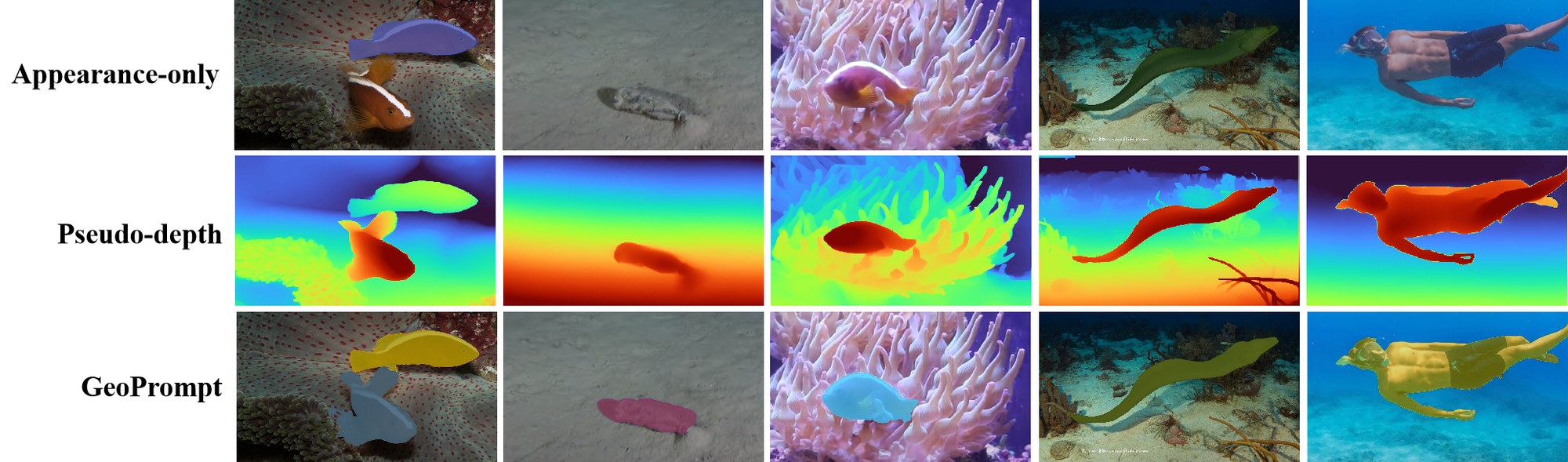}
    \caption{
Visual comparison of appearance-only predictions, pseudo-depth maps, and GeoPrompt predictions.
Appearance-only predictions are prone to incomplete masks or inaccurate localization, while pseudo-depth maps reveal object contours and spatial layout.
By leveraging depth-derived geometry, GeoPrompt produces more complete and reliable open-vocabulary instance predictions.
}

\label{fig:motivation}
\end{figure*}

\section{Introduction}

Open-vocabulary instance segmentation (OVIS) aims to segment object instances from both seen and unseen categories by aligning visual region representations with textual category embeddings~\cite{clip,lseg,maskclip_ovs,ovseg,odise,san,freeseg,fcclip}. Benefiting from vision-language models, OVIS reduces the dependence on closed-set category annotations and provides a promising paradigm for marine perception, where underwater environments contain diverse, fine-grained, and long-tail object categories. This ability is especially important for underwater scene understanding, since many marine objects are difficult to predefine and annotate in practical applications~\cite{maris}.

Despite recent progress, underwater open-vocabulary instance segmentation remains highly challenging. Most existing OVIS methods are developed under the assumption that RGB visual appearance can provide discriminative cues for matching textual category embeddings~\cite{catseg,scan,sed,image_embedding_balancing,maftplus,eovseg}. However, underwater imaging conditions often violate this assumption. Color attenuation, light scattering, turbidity, low contrast, and non-uniform illumination distort object appearance~\cite{uimage_formation,suim,watermask,usis_sam,diveseg}, causing visual region features to deviate from their corresponding semantic concepts. Since open-vocabulary recognition depends on comparing visual features with textual classifiers, such appearance drift directly weakens vision-language alignment, especially for unseen categories that rely more heavily on transferable semantic correspondence.

To address this challenge, MARIS~\cite{maris} introduces a marine OVIS benchmark and improves underwater segmentation through geometric enhancement and semantic alignment. Nevertheless, existing methods still leave an important problem underexplored: underwater degradation affects not only the visual representation, but also the way visual regions should be semantically aligned with textual classifiers.

Prompt-based learning has become an effective strategy for adapting vision-language models to downstream tasks by replacing hand-crafted templates with learnable or input-conditioned textual contexts~\cite{coop,cocoop,maple,promptsrc}. In open-vocabulary segmentation, such prompts are used to construct textual classifiers, which directly influence the matching between visual regions and category semantics~\cite{clip,ovseg,fcclip}. Fixed prompts preserve general semantic knowledge but cannot adapt to scene-specific underwater degradation. Image-conditioned prompt learning can generate adaptive prompts from visual features~\cite{cocoop}, yet the image features used for prompt generation may also be affected by underwater appearance drift. Therefore, semantic adaptation based only on fixed templates or degraded RGB features remains insufficient for robust underwater open-vocabulary recognition.

Geometry provides a natural complementary cue to alleviate this problem. Prior RGB-D, monocular depth estimation, and depth-aware segmentation studies have shown that geometric information can complement RGB appearance by providing object boundaries, spatial layout, and shape cues for dense prediction tasks~\cite{fusenet,rdfnet,rednet,sagate,dpt,depth_anything,depth_anything_v2}. Inspired by this observation, we use depth-derived intermediate features as structural priors for underwater perception. Importantly, our method does not require ground-truth depth annotations or explicit depth supervision. Instead of treating monocular depth as a target prediction, we exploit depth-derived features to provide geometry-aware cues that are complementary to degraded RGB appearance. Figure~\ref{fig:motivation} visualizes pseudo-depth maps as intermediate geometry cues, showing that they provide useful structural information for improving underwater open-vocabulary instance prediction.

Based on this insight, we propose \textbf{GeoPrompt} for underwater open-vocabulary instance segmentation, which introduces geometry-conditioned prompt learning to improve vision-language alignment. GeoPrompt adapts both sides of vision-language alignment. On the visual side, a geometry-guided visual adaptation module extracts depth-derived structural features and injects them into multi-scale visual representations, helping the segmentation decoder capture more stable object structure under underwater degradation. On the semantic side, a geometry-conditioned prompt branch generates geometry-aware textual classifiers from structural representations. These classifiers are integrated with the original fixed classifier and the image-conditioned prompt classifier, allowing the semantic space to adapt not only to image appearance but also to underwater geometry.
By introducing geometry-guided visual adaptation and geometry-conditioned prompt learning, GeoPrompt improves both visual representation and semantic alignment for seen and unseen underwater categories.

Our contributions are summarized as follows:

\begin{itemize}
    \item We reformulate underwater open-vocabulary instance segmentation as a geometry-conditioned vision-language alignment problem, emphasizing that both visual representations and semantic prompts should be adapted under underwater degradation.

    \item We propose GeoPrompt, a unified framework consisting of Geometry-Guided Visual Adaptation Module (GVAM) and Geometry-Conditioned Prompt Learning (GCPL). The former injects depth-derived structural priors into multi-scale visual features, while the latter uses geometric cues to generate degradation-aware semantic prompts.

    \item We design a \textbf{Geometry-Conditioned Vision-Language Alignment} strategy that aligns geometry-enhanced visual embeddings with geometry-aware textual classifiers for open-vocabulary mask prediction. Experiments on MARIS show that GeoPrompt achieves state-of-the-art performance and improves generalization to unseen underwater categories.
\end{itemize}

\begin{figure*}[!t]
    \centering
    \includegraphics[width=0.95\textwidth]{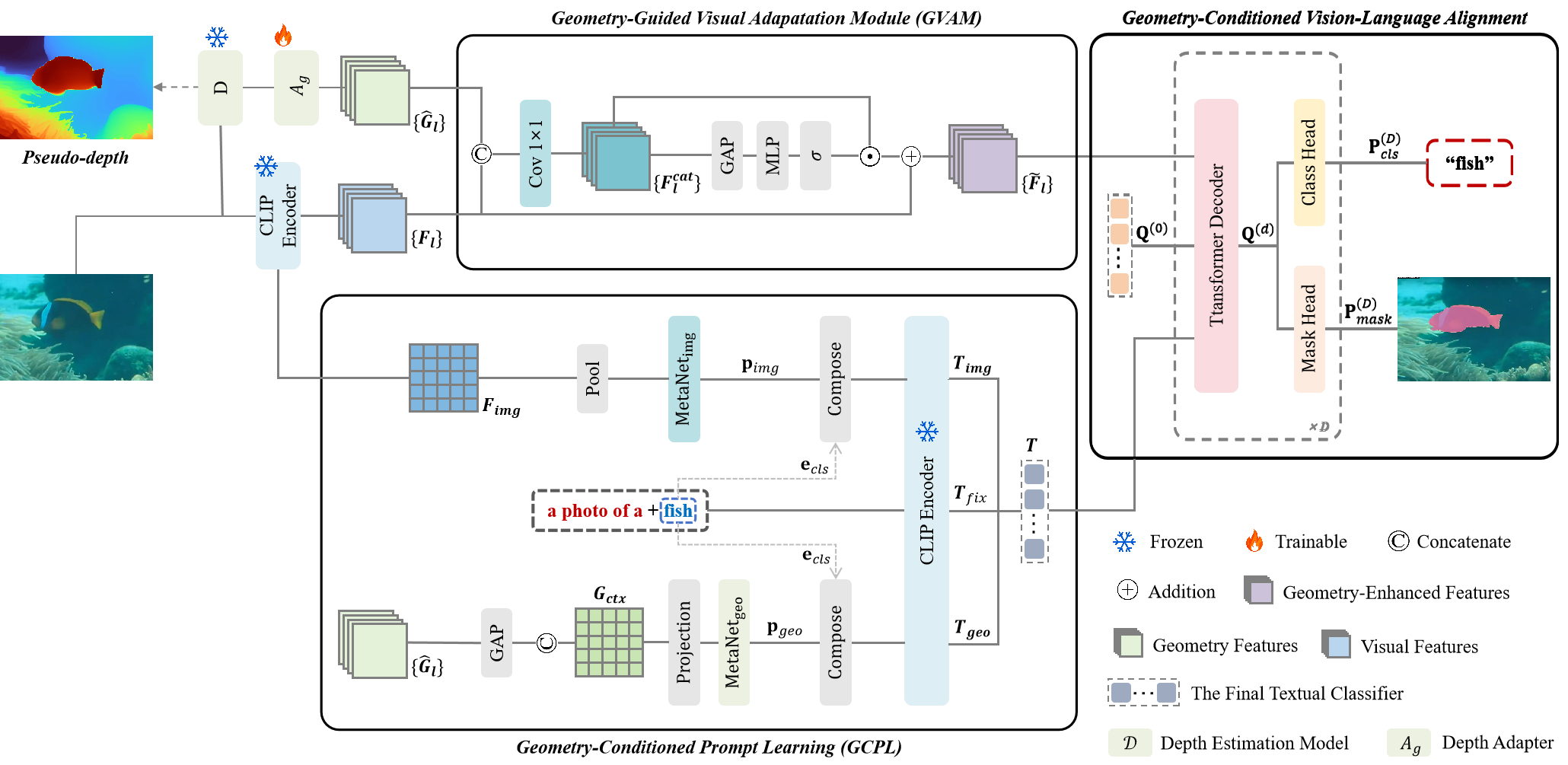}
    \caption{
    Overall architecture of GeoPrompt.
GeoPrompt consists of GVAM and GCPL.
GVAM extracts depth-derived structural features from a frozen depth estimation model and adapts them to obtain geometry-enhanced visual representations, with the pseudo-depth map shown only as a visualization of the geometry cue.
GCPL generates geometry-aware textual classifiers and integrates them with fixed and image-conditioned classifiers for geometry-conditioned vision-language alignment and open-vocabulary instance mask prediction.
    }
    \label{fig:geoprompt_pipeline}
\end{figure*}

\section{Related Work}

\subsection{Open-Vocabulary Instance Segmentation}

Open-vocabulary instance segmentation (OVIS) aims to recognize and segment object instances from both seen and unseen categories by leveraging the semantic generalization of vision-language models~\cite{clip,ovseg,san}. Existing approaches often decouple mask generation from category prediction: a segmentation decoder produces instance masks, while visual region embeddings are matched with textual classifiers derived from category names or prompt templates~\cite{ovseg,odise,fcclip}. This design enables recognition beyond the predefined training label space. However, most methods are developed on natural-image benchmarks and assume that RGB features provide reliable cues for vision-language alignment.

This assumption is particularly fragile in underwater environments. Color attenuation, scattering, turbidity, and low contrast distort visual appearance, causing region features to deviate from their corresponding textual concepts. Related semantic segmentation studies have addressed distribution shifts through curriculum-style self-training for source-free adaptation~\cite{wang2024curriculum} and informative data mining for one-shot cross-domain adaptation~\cite{wang2023informative}. Although these methods do not directly target underwater OVIS, they highlight the importance of robust representations under distribution shifts. Furthermore, geometry-guided self-distillation has been explored for open-vocabulary 3D scene understanding~\cite{wang2024geometry}, demonstrating the potential of structural cues for open-vocabulary representations.

MARIS~\cite{maris} takes an important step toward underwater OVIS by introducing a marine benchmark and enhancing visual representations through geometric features and semantic alignment. Nevertheless, existing underwater OVIS approaches primarily employ geometry for visual feature enhancement, while geometry-conditioned adaptation of textual classifiers remains underexplored. This motivates GeoPrompt, which uses depth-derived structural representations to guide both visual adaptation and textual classifier generation for more reliable open-vocabulary alignment.

\subsection{Geometry-Conditioned Prompt Learning}

Prompt learning adapts vision-language models by replacing handcrafted textual templates with learnable or input-conditioned context tokens. CoOp~\cite{coop} learns task-specific context vectors, while CoCoOp~\cite{cocoop} introduces image-conditioned prompts to improve generalization to unseen categories. MaPLe~\cite{maple} further explores multimodal prompt learning by jointly adapting visual and textual representations. These approaches provide flexible textual classifiers, but their effectiveness depends on the reliability of the features used to generate adaptive prompts.

In underwater scenes, image-conditioned prompts may inherit appearance bias from degraded RGB features. Color distortion, scattering, and background interference can affect both visual region embeddings and image-level features used for prompt generation, weakening semantic alignment. Geometry offers a complementary cue, since structural information such as object contours, shapes, and spatial layout can remain informative under appearance degradation. Previous RGB-D segmentation studies~\cite{fusenet,sagate} have demonstrated the benefits of incorporating geometric cues into visual representations. However, these methods primarily exploit geometry for visual feature enhancement rather than textual classifier adaptation.

To address this limitation, GeoPrompt introduces geometry-conditioned prompt learning that generates adaptive textual classifiers from depth-derived structural features. By integrating geometry-aware prompts with fixed and image-conditioned classifiers, GeoPrompt adapts both sides of vision-language alignment, reducing dependence on degraded RGB appearance and improving semantic correspondence for underwater open-vocabulary recognition.

\section{Method}

The overall architecture of GeoPrompt is shown in Figure~\ref{fig:geoprompt_pipeline}. It introduces three core components: (1) a \textbf{Geometry-Guided Visual Adaptation Module (GVAM)} that extracts depth-derived structural priors and injects them into multi-scale visual features, alleviating underwater appearance drift; (2) a \textbf{Geometry-Conditioned Prompt Learning (GCPL)} module that generates geometry-aware textual classifiers from structural representations, reducing semantic misalignment; (3) a \textbf{Geometry-Conditioned Vision-Language Alignment} mechanism that matches geometry-enhanced visual features with adaptive textual classifiers for robust open-vocabulary mask prediction. Through this design, GeoPrompt adapts both sides of vision-language alignment, enabling more reliable recognition of both seen and unseen underwater categories.

\subsection{Geometry-Guided Visual Adaptation}

Underwater images are often degraded by color attenuation, scattering, turbidity, and low contrast, which weaken RGB-based representations for open-vocabulary instance segmentation~\cite{watermask,usis_sam,diveseg}. To complement unreliable appearance cues, geometric structures such as object contours, shapes, spatial layout, and depth-related hierarchy provide useful structural information for dense prediction~\cite{fusenet,rdfnet,rednet,sagate}. We therefore introduce a Geometry-Guided Visual Adaptation Module that extracts depth-derived structural features to enhance multi-scale visual representations.

\textbf{Geometry Feature Extraction.}
Given an input image $x \in \mathbb{R}^{H \times W \times 3}$, we first extract multi-scale visual features $\{F_l\}_{l=1}^{L}$ from the CLIP image encoder~\cite{clip}, where $L$ denotes the number of feature scales. In parallel, a pre-trained monocular depth estimation model $\mathcal{D}$, implemented with Depth Anything V2~\cite{depth_anything_v2}, is employed as a geometry encoder to produce depth-derived structural features. Specifically, we extract intermediate features from the depth encoder at multiple stages:
\begin{equation}
\{G_l\}_{l=1}^{L} = \mathcal{D}_{\text{intermediate}}(x),
\end{equation}
where $G_l \in \mathbb{R}^{C_g \times H_l \times W_l}$ represents the geometry feature at scale $l$.

Since the raw geometry features are not directly optimized for underwater instance segmentation, we further introduce a lightweight geometry adapter to transform them into task-specific structural representations:
\begin{equation}
\hat{G}_l = A_g(G_l),
\end{equation}
where $A_g(\cdot)$ denotes the geometry adapter and $\hat{G}_l$ denotes the adapted geometry feature at scale $l$. These adapted geometry features provide structural cues that are complementary to degraded RGB appearance.

\textbf{Visual-Geometry Fusion.}
After obtaining the adapted geometry features $\hat{G}_l$, we fuse them with the corresponding visual features $F_l$ at each scale. Following the geometry-enhanced fusion paradigm in marine open-vocabulary instance segmentation~\cite{maris}, we adopt a Visual-Geometric Multimodal Interaction (VGMI) block for cross-modal fusion. The VGMI block performs feature concatenation, channel transformation, and channel-wise gating:
\begin{equation}
F_l^{\text{cat}} = \text{Conv}_{1\times1}\left(\text{Concat}(F_l, \hat{G}_l)\right),
\end{equation}
\begin{equation}
\tilde{F}_l =
\sigma\left(\text{MLP}\left(\text{GAP}(F_l^{\text{cat}})\right)\right)
\odot F_l^{\text{cat}} + F_l ,
\end{equation}
where $\text{GAP}$ denotes global average pooling, $\sigma$ is the sigmoid function, and $\odot$ denotes element-wise multiplication. The channel gate selectively emphasizes informative feature channels, while the residual connection preserves the original visual semantics. The output $\tilde{F}_l$ denotes the geometry-enhanced visual feature at scale $l$.

This fusion strategy allows the model to exploit both RGB appearance information and geometry priors. It is particularly useful in underwater scenes, where object texture and color can be severely degraded while structural cues such as contours and spatial layout remain relatively stable.

\subsection{Geometry-Conditioned Prompt Learning}

Open-vocabulary instance segmentation relies on textual classifiers generated from category names and prompt templates~\cite{clip,ovseg,odise,san,fcclip}. Fixed prompts such as ``a photo of a [CLASS]'' preserve general semantic knowledge from pre-training~\cite{clip,coop} but cannot adapt to scene-specific underwater degradation. Image-conditioned prompt learning further generates adaptive prompts from image features~\cite{cocoop}:
\begin{equation}
\mathbf{p}_{\text{img}} =
\text{MetaNet}_{\text{img}}\left(\text{Pool}(F_{\text{img}})\right),
\end{equation}
where $F_{\text{img}}$ denotes the image-level visual feature and $\text{Pool}(\cdot)$ denotes global feature aggregation, such as GAP or a CLS token.

However, in underwater scenes, the image features used for prompt generation may themselves be affected by color attenuation, scattering, and low contrast~\cite{watermask,usis_sam,diveseg}. As a result, image-conditioned prompts may inherit degraded appearance bias, leading to semantic misalignment between visual regions and textual classifiers. To address this problem, we introduce a geometry-conditioned prompt branch that generates textual classifiers from depth-derived structural representations.

\textbf{Geometry Context Aggregation.}
We first aggregate the adapted multi-scale geometry features into a global geometry context:
\begin{equation}
G_{\text{ctx}} =
\text{Concat}
\left(
\text{GAP}(\hat{G}_1),
\text{GAP}(\hat{G}_2),
\ldots,
\text{GAP}(\hat{G}_L)
\right).
\end{equation}
The concatenated context $G_{\text{ctx}}$ captures multi-scale geometric structure information that is complementary to RGB appearance~\cite{fusenet,rdfnet,rednet,sagate}.

\textbf{Geometry-Conditioned Prompt Generation.}
The geometry context $G_{\text{ctx}}$ is fed into a Geometry Prompt Adapter to generate geometry-conditioned prompt tokens. Specifically, we first project the geometry context into a compact representation:
\begin{equation}
G_{\text{proj}} =
\text{ReLU}
\left(
\text{Linear}
\left(
\text{LayerNorm}(G_{\text{ctx}})
\right)
\right),
\end{equation}
and then generate geometry-conditioned prompt tokens:
\begin{equation}
\mathbf{p}_{\text{geo}} =
\text{MetaNet}_{\text{geo}}(G_{\text{proj}}).
\end{equation}
These geometry-conditioned prompt tokens are combined with category name embeddings and encoded by the frozen CLIP text encoder~\cite{clip} to produce a geometry-aware textual classifier:
\begin{equation}
\mathbf{T}_{\text{geo}} =
\text{CLIP}_{\text{text}}
\left(
\text{compose}(\mathbf{p}_{\text{geo}}, \mathbf{e}_{\text{cls}})
\right),
\end{equation}
where $\mathbf{e}_{\text{cls}}$ denotes the CLIP text embedding of the category name.

\textbf{Multi-Source Classifier Integration.}
GeoPrompt integrates three complementary textual classifiers: the original fixed classifier, the image-conditioned classifier~\cite{cocoop}, and the geometry-conditioned classifier. The final textual classifier is computed as:
\begin{equation}
\mathbf{T} =
\text{Norm}
\left(
(1-\lambda-\alpha_g)\mathbf{T}_{\text{fix}}
+
\lambda\mathbf{T}_{\text{img}}
+
\alpha_g\mathbf{T}_{\text{geo}}
\right),
\end{equation}
where $\mathbf{T}_{\text{fix}}$ denotes the fixed textual classifier, $\mathbf{T}_{\text{img}}$ denotes the image-conditioned classifier, and $\mathbf{T}_{\text{geo}}$ denotes the geometry-conditioned classifier. The coefficient $\lambda$ controls the contribution of the image-conditioned branch, while $\alpha_g$ controls the contribution of the geometry-conditioned branch.

For ablation studies, this formulation supports flexible configurations. Removing the image-conditioned branch yields a Fixed+Geo variant:
\begin{equation}
\mathbf{T} =
\text{Norm}
\left(
(1-\alpha_g)\mathbf{T}_{\text{fix}}
+
\alpha_g\mathbf{T}_{\text{geo}}
\right).
\end{equation}
Removing the geometry-conditioned branch recovers the fixed-plus-image-conditioned prompt variant.

\begin{table*}[!t]
\centering
\caption{Comparison with the state-of-the-art open-vocabulary segmentation methods on the MARIS benchmark.The results of OVSeg, ODISE, SAN, FCCLIP, MAFT+, EOVSeg, and MARIS are reported by the MARIS benchmark. GeoPrompt is evaluated under the same in-domain setting.}
\label{tab:sota}
\resizebox{\textwidth}{!}{
\begin{tabular}{lllccccccccc}
\toprule
Method & Publication & Backbone
& \multicolumn{3}{c}{Intersection Class}
& \multicolumn{3}{c}{Open-Vocabulary Class}
& \multicolumn{3}{c}{Overall Class} \\
\cmidrule(lr){4-6} \cmidrule(lr){7-9} \cmidrule(lr){10-12}
& & 
& mAP & AP$_{50}$ & AP$_{75}$
& mAP & AP$_{50}$ & AP$_{75}$
& mAP & AP$_{50}$ & AP$_{75}$ \\
\midrule
OVSeg  & CVPR'23   & ViT-B
& 37.52 & 48.51 & 43.26
& 27.21 & 33.65 & 30.38
& 30.95 & 39.02 & 35.47 \\

ODISE  & CVPR'23   & ViT-B
& 41.89 & 50.74 & 46.83
& 30.26 & 35.68 & 32.54
& 34.71 & 41.56 & 38.12 \\

SAN    & CVPR'23   & ViT-B
& 43.26 & 52.18 & 48.05
& 31.57 & 37.09 & 34.02
& 36.05 & 43.06 & 39.26 \\

FCCLIP & NeurIPS'23 & ConvNeXt-B
& 47.78 & 57.22 & 52.44
& 34.53 & 39.84 & 37.15
& 39.26 & 46.03 & 42.60 \\

MAFT+  & ECCV'24   & ConvNeXt-B
& 48.15 & 58.26 & 54.57
& 35.72 & 40.67 & 38.88
& 40.08 & 47.16 & 43.33 \\

EOVSeg & AAAI'25  & ConvNeXt-B
& 37.98 & 48.95 & 41.55
& 27.48 & 33.89 & 29.56
& 31.22 & 39.26 & 33.83 \\

MARIS  & CVPR'26  & ConvNeXt-B
& 52.68 & 61.56 & 57.33
& 39.77 & 45.78 & 42.68
& 44.37 & 51.41 & 47.90 \\

\textbf{GeoPrompt (Ours)}
& -- & ConvNeXt-B
& \textbf{56.34} & \textbf{65.75} & \textbf{60.88}
& \textbf{48.20} & \textbf{54.88} & \textbf{51.49}
& \textbf{51.10} & \textbf{58.76} & \textbf{54.84} \\
\bottomrule
\end{tabular}
}
\end{table*}

\subsection{Geometry-Conditioned Vision-Language Alignment}

After obtaining the geometry-enhanced visual features $\{\tilde{F}_l\}_{l=1}^{L}$ and the final textual classifier $\mathbf{T}$, the segmentation decoder predicts instance masks and assigns open-vocabulary category labels by matching visual mask embeddings with textual classifiers~\cite{clip,ovseg,fcclip}. This process performs geometry-conditioned vision-language alignment from both sides: the visual features are enhanced by depth-derived structural priors, and the textual classifier is adapted by geometry-conditioned prompt learning.

Following the MARIS framework~\cite{maris}, we employ a Mask2Former-style transformer decoder~\cite{maskformer,mask2former}. The decoder takes multi-scale visual features as input and refines a set of learnable query embeddings through multiple decoder layers. At each decoder layer, the model predicts mask embeddings and classification embeddings:
\begin{equation}
\mathbf{M}^{(d)} =
\text{MLP}_{\text{mask}}(\mathbf{Q}^{(d)}),
\quad
\mathbf{C}^{(d)} =
\text{MLP}_{\text{cls}}(\mathbf{Q}^{(d)}),
\end{equation}
where $d$ indexes the decoder layer. The mask predictions are obtained by computing the similarity between mask embeddings and pixel-level visual features:
\begin{equation}
\mathbf{P}_{\text{mask}}^{(d)}
=
\mathbf{M}^{(d)}
\cdot
\tilde{\mathbf{F}}_{\text{px}}^{\top}.
\end{equation}

To perform open-vocabulary classification, instance-level visual embeddings are compared with the final textual classifier~\cite{clip,ovseg,fcclip}:
\begin{equation}
\mathbf{P}_{\text{cls}}^{(d)}
=
\gamma_{\text{logit}}
\cdot
\frac{\mathbf{C}^{(d)}}{\|\mathbf{C}^{(d)}\|_2}
\cdot
\frac{\mathbf{T}^{\top}}{\|\mathbf{T}\|_2},
\end{equation}
where $\gamma_{\text{logit}}$ is a learnable logit scale. The final classifier $\mathbf{T}$ incorporates fixed semantic knowledge, image-conditioned adaptation, and geometry-conditioned semantic cues, enabling more robust category prediction under underwater degradation.

\textbf{Training Objective.}
GeoPrompt is optimized using the standard open-vocabulary instance segmentation objective~\cite{mask2former,maris}:
\begin{equation}
\mathcal{L}
=
\mathcal{L}_{\text{cls}}
+
\mathcal{L}_{\text{mask}}
+
\mathcal{L}_{\text{dice}}.
\end{equation}
Here, $\mathcal{L}_{\text{cls}}$ is the classification loss, $\mathcal{L}_{\text{mask}}$ is the mask loss, and $\mathcal{L}_{\text{dice}}$ is the DICE loss. The Hungarian matching algorithm is used for bipartite matching between predictions and ground truth, with deep supervision applied to auxiliary decoder outputs~\cite{mask2former}.

\section{Experiments}

\subsection{Experimental Setup}
\paragraph{Dataset.}
We conduct experiments on the MARIS-OpenIS benchmark for underwater open-vocabulary instance segmentation. MARIS-OpenIS contains more than 16K underwater images with pixel-level instance masks, covering 9 super-classes and 158 fine-grained subclasses. It refines coarse underwater labels such as ``fish'' and ``plants'' into more detailed marine categories, making it suitable for evaluating fine-grained open-vocabulary recognition. Following the official protocol, the dataset is split into 5,712 training images and 10,439 validation images. The training and validation sets contain 84 and 115 categories, respectively, with 41 overlapping categories. We report results on both intersection categories and open-vocabulary categories that appear only in the validation set.

\paragraph{Evaluation Metrics.}
We report COCO-style mask AP~\cite{coco}, including AP, AP$_{50}$, AP$_{75}$. When category splits are available, we further report AP on seen/intersection and open-vocabulary categories.

\paragraph{Implementation Details.}
We use ConvNeXt-Base~\cite{convnext} CLIP~\cite{clip} as the vision-language backbone and Depth Anything V2~\cite{depth_anything_v2} with ViT-B as the geometry prior extractor. The model is trained for 50K iterations with a total batch size of 16 on 8 GPUs. The base learning rate is set to $1\times10^{-4}$. For prompt learning, the number of context tokens is set to 4 and initialized with ``a photo of a''. The image-conditioned prompt coefficient is set to $\lambda=0.05$, while the geometry-conditioned prompt coefficient is initialized as $\alpha_g=0.05$ and constrained by a maximum value of 0.10.

\subsection{Comparison with the State-of-the-Art Methods}
\subsubsection{Quantitative Results}

Table~\ref{tab:sota} compares GeoPrompt with representative open-vocabulary segmentation methods on the MARIS benchmark. We compare with existing methods including OVSeg~\citep{ovseg}, ODISE~\citep{odise}, SAN~\citep{san}, FCCLIP~\citep{fcclip}, MAFT+~\citep{maftplus}, EOVSeg~\citep{eovseg}, and the marine open-vocabulary baseline MARIS~\citep{maris}. Existing results are taken from the MARIS benchmark~\citep{maris}, and our GeoPrompt is evaluated under the same in-domain setting. Compared with previous methods, GeoPrompt achieves the best performance on intersection, open-vocabulary, and overall categories, demonstrating the effectiveness of geometry-conditioned prompt learning for underwater open-vocabulary instance segmentation.

Compared with the MARIS baseline, GeoPrompt improves the overall mAP from 44.37 to 51.10, bringing a gain of 6.73 points. On open-vocabulary categories, GeoPrompt improves mAP from 39.77 to 48.20, with a gain of 8.43 points. This improvement is especially important because open-vocabulary categories rely more heavily on transferable vision-language alignment rather than category-specific supervision. These results show that geometry-conditioned prompt learning and geometry-guided visual adaptation can effectively reduce the negative influence of underwater appearance degradation and improve the recognition of unseen underwater objects.

\begin{figure}[!t]
    \centering
    \includegraphics[width=0.99\linewidth]{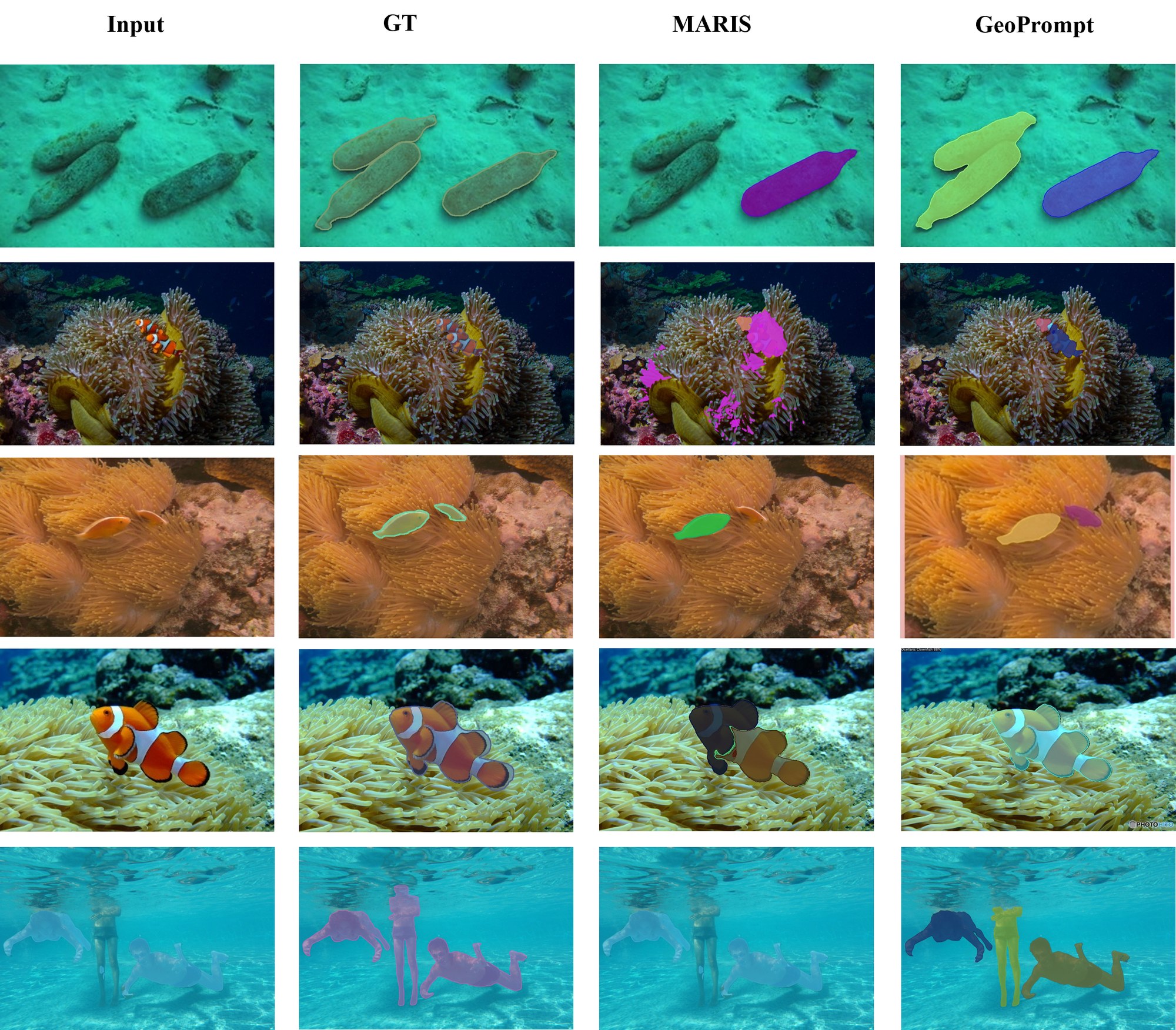}
    \caption{
    Qualitative comparison between MARIS and GeoPrompt.
    From left to right: input image, ground-truth annotation, prediction of MARIS, and prediction of GeoPrompt.
    Compared with MARIS, GeoPrompt produces more complete masks, clearer object boundaries, and more accurate instance-level predictions.
    }
    \label{fig:qualitative_results}
\end{figure}

\begin{table*}[t]
\centering
\caption{Ablation study on the main components of GeoPrompt.}
\label{tab:main_ablation}
\small
\renewcommand{\arraystretch}{1.08}
\setlength{\tabcolsep}{3.5pt}

\begin{tabularx}{0.98\textwidth}{ccYYYYYYYYY}
\toprule
\multirow{2}{*}{GVAM}
& \multirow{2}{*}{GCPL}
& \multicolumn{3}{c}{Intersection Class}
& \multicolumn{3}{c}{Open-Vocabulary Class}
& \multicolumn{3}{c}{Overall Class} \\
\cmidrule(lr){3-5}
\cmidrule(lr){6-8}
\cmidrule(lr){9-11}
&
& mAP & AP$_{50}$ & AP$_{75}$
& mAP & AP$_{50}$ & AP$_{75}$
& mAP & AP$_{50}$ & AP$_{75}$ \\
\midrule

$\times$ & $\times$
& 52.68 & 61.56 & 57.33
& 39.77 & 45.78 & 42.68
& 44.37 & 51.41 & 47.90 \\

$\checkmark$ & $\times$
& 55.09 & 64.56 & 60.02
& 48.46 & 55.55 & 52.02
& 50.83 & 58.76 & 54.87 \\

$\times$ & $\checkmark$
& 49.13 & 57.64 & 53.15
& 45.12 & 51.93 & 48.51
& 46.55 & 53.97 & 50.16 \\

\rowcolor{gray!15}
$\checkmark$ & $\checkmark$
& \textbf{56.34} & \textbf{65.75} & \textbf{60.88}
& 48.20 & 54.88 & 51.49
& \textbf{51.10} & \textbf{58.76} & 54.84 \\

\bottomrule
\end{tabularx}

\end{table*}

\begin{figure*}[!t]
    \centering
    \includegraphics[width=0.98\textwidth]{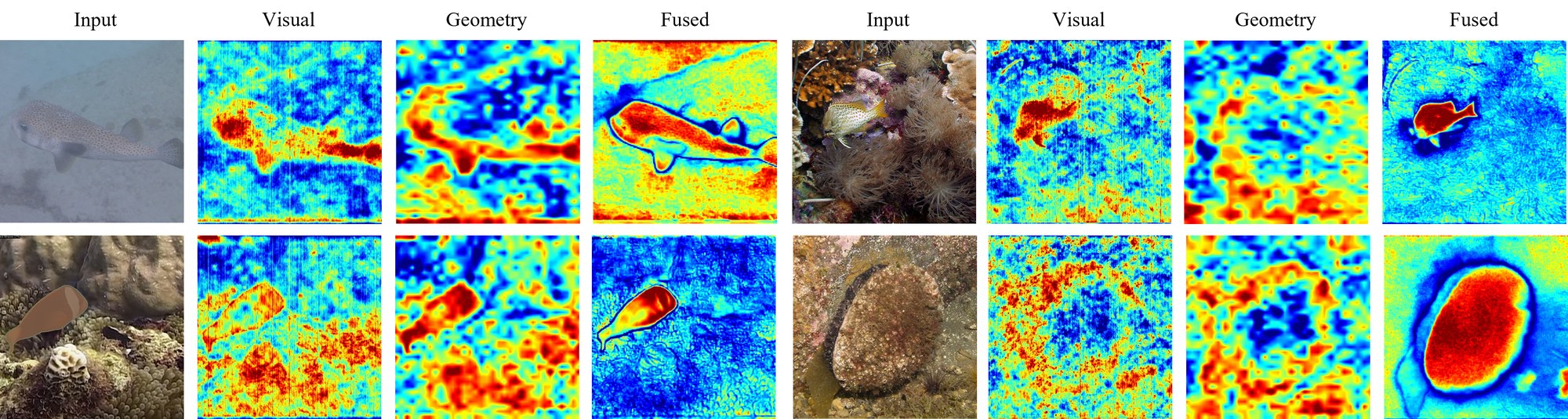}
    \caption{
Qualitative visualization of image-side feature responses.
Each group shows the input image, visual response, geometry response, and fused response.
The fused response becomes more concentrated on object-related regions after visual-geometric adaptation.
}
    \label{fig:feature_heatmap}
\end{figure*}

\subsubsection{Qualitative Results}

Figure~\ref{fig:qualitative_results} presents qualitative comparisons between GeoPrompt and MARIS, the strongest baseline in our quantitative comparison and the most relevant geometry-enhanced underwater OVIS method. Compared with MARIS, GeoPrompt produces more complete and accurate mask predictions in challenging scenes. In the first row, GeoPrompt successfully identifies multiple object instances, whereas MARIS misses part of the targets. In the second and third rows, GeoPrompt yields clearer object boundaries and more reliable localization for small or partially ambiguous instances. In the fourth row, GeoPrompt better preserves the object shape and extent, showing improved structural consistency. In the last row, GeoPrompt separates adjacent instances more clearly, while MARIS tends to produce incomplete or inaccurate masks. Overall, these qualitative results show that GeoPrompt provides more reliable instance localization, finer boundary delineation, and stronger category discrimination than MARIS, demonstrating the benefit of jointly exploiting geometry-guided visual adaptation and geometry-aware prompt learning.
\subsection{Ablation Experiments}

\paragraph{Main Component Ablation.} 
To evaluate the contribution of the two main components, we conduct ablation experiments on GVAM and GCPL. GVAM introduces geometry-guided visual adaptation to enhance multi-scale visual representations, while GCPL introduces geometry-conditioned prompt learning to adapt textual classifiers with structural context.

As shown in Table~\ref{tab:main_ablation}, the baseline without GVAM and GCPL obtains 44.37 overall mAP. Introducing GVAM improves the overall mAP to 50.83, showing that geometry-guided visual adaptation provides effective structural cues for underwater instance segmentation. Introducing GCPL alone improves the overall mAP to 46.55, indicating that geometry-conditioned prompt learning can improve semantic alignment even without visual-side geometry fusion.

When GVAM and GCPL are combined, GeoPrompt achieves the best overall mAP of 51.10 and the best intersection-class mAP of 56.34. These results demonstrate that visual-side geometry adaptation and prompt-side geometry-conditioned classifier generation are complementary. Although GVAM alone already provides strong improvements on open-vocabulary categories, the full model achieves the best overall balance across intersection, open-vocabulary, and overall categories.

\paragraph{Feature Response Visualization.}
To further analyze the effect of geometry-guided visual adaptation, we visualize the image-side feature responses of GeoPrompt in Figure~\ref{fig:feature_heatmap}. Each row shows the input image, visual response, geometry response, and fused response. The visual response is obtained from RGB visual features, while the geometry response is derived from depth-based structural representations after geometry adaptation. The fused response represents the geometry-enhanced visual representation after visual-geometric adaptation.

As shown in Figure~\ref{fig:feature_heatmap}, the visual response is often affected by underwater appearance degradation, such as color attenuation, low contrast, and background clutter. In contrast, the geometry response highlights more stable structural regions, including object contours and spatial layouts. After visual-geometric fusion, the fused response becomes more concentrated on object regions and shows clearer structural activation. These qualitative results indicate that GVAM helps the model exploit geometry cues to obtain more robust image-side representations under challenging underwater conditions.

\section{Conclusion}
In this paper, we proposed GeoPrompt, a geometry-conditioned prompt learning framework for underwater open-vocabulary instance segmentation. GeoPrompt consists of two complementary components: GVAM and GCPL. GVAM enhances multi-scale visual features with depth-derived geometric representations, while GCPL generates adaptive textual classifiers from image-level and geometry-level contexts. By jointly adapting visual representations and textual classifiers, GeoPrompt improves geometry-conditioned vision-language alignment under underwater appearance degradation. Importantly, GeoPrompt does not require ground-truth depth annotations, making it practical for RGB-only marine datasets. Experiments on the MARIS benchmark demonstrate the effectiveness of GeoPrompt. GeoPrompt improves OV mAP from 39.77 to 48.20, showing stronger generalization to unseen underwater categories. Ablation studies and qualitative results further validate the contribution of the proposed components.

Despite these improvements, challenging cases remain. Extremely small objects, heavy occlusion, ambiguous boundaries, or noisy depth-derived geometry features may still lead to incomplete predictions. Future work will explore robust geometry priors and stronger semantic discrimination for complex marine scenarios.

\section*{Acknowledgments}

The authors would like to thank their colleagues for
their valuable discussions and constructive suggestions
during this research.

\bibliography{aaai2027}
\end{document}